%% file: smt-arxiv-corrected.tex
\def\year{2020}\relax
\documentclass[letterpaper]{article} 
\usepackage{aaai20}  
\usepackage{amsmath}
\usepackage{amssymb}
\usepackage{graphicx,txfonts}
\usepackage{pifont}
\usepackage{fancyvrb}
\usepackage{caption}
\usepackage{subcaption}
\usepackage{pdfpages}
\usepackage{float}
\usepackage{stmaryrd,verbatim}

\input defnj  

\begin{document}


\title{SMT + ILP\thanks{The author was supported by a Royal Society University Research Fellowship.}
}
\author{Vaishak Belle \\ 
University of Edinburgh \& Alan Turing Institute \\ 
\small  vaishak@ed.ac.uk}

\maketitle
\begin{abstract} \small Inductive logic programming (ILP) has been a deeply influential paradigm in AI, enjoying decades of research on its theory and implementations. As a natural descendent of the fields of logic programming and machine learning, it admits the incorporation of  background knowledge, which can be very useful in domains where prior knowledge from experts is available and can lead to a more data-efficient learning regime. 

Be that as it may, the limitation to Horn clauses composed over Boolean variables is a very serious one. Many phenomena occurring in the real-world  are best characterized using continuous entities, and more generally, mixtures of discrete and continuous entities. In this position paper, we motivate a reconsideration of inductive declarative programming by leveraging satisfiability modulo theory technology.

\end{abstract}

\section{Introduction}

Inductive logic programming (ILP) has been a deeply influential paradigm in AI, enjoying decades of research on its theory and implementations \cite{inductive-logic-programming:-theory,probabilistic-inductive-logic-programming:,muggleton2012ilp}. ILP continues to be applied in domains ranging from robotics to biology. As a natural descendent of the fields of logic programming and machine learning, it admits the incorporation of complex background knowledge, which can be very useful in domains where prior knowledge from experts is available and can lead to a more data-efficient learning regime. In essence, the semantic theory attempts to construct a hypothesis based on entailment judgements wrt a (possibly small) set of examples. The construction itself may appeal to principled notions such as inverse entailment \cite{Muggleton:1991:ILP:104992.104994}. 

Be that as it may, the limitation to Horn clauses composed over Boolean variables is a very serious one. Many phenomena occurring in the real-world, from gravitational and quantum mechanics to stock price fluctuations, are best characterized using continuous entities, and more generally, mixtures of discrete and continuous entities. While many notable proposals are treating the issue of modeling hybrid phenomena in logic programming and inductive logic programming settings  \cite{hybrid-probabilistic-logic-programming,speichert2018learning},  the underpinning semantics still largely reduces to classical notions with continuous concepts carefully (but not generally) integrated.  Thus, we consider whether we should upgrade the logical basis for ILP to natively handle continuous concepts.

In recent years, \emph{satisfiability modulo theories}    (SMT) has emerged as a pragmatic logical framework for reasoning about complex terms, inequalities and other arithmetic operations \cite{satisfiability-modulo-theories}, such as testing the satisfiability of linear constraints. 
For example, in \cite{approximate-counting-in-smt-and-value-estimation,probabilistic-inference-in-hybrid-domains}, SMT solvers were used to generalize model counting to hybrid domains by computing the volume of the polytope encoded as a linear constraint over the reals. Other  significant advances in SMT have included the handling of inductive constraints and non-linear theories  \cite{reynolds2015induction,dreal:-an-smt-solver-for-nonlinear-theories}. Thus, we argue that SMT and similar technologies could serve as a reasonable basis to upgrade ILP for hybrid domains. We do not suggest the proposal should aim to subsume classical ILP, as it is very likely that Horn logic over Boolean atoms will be both sufficient and efficient for many problem domains. Rather, it is meant to be complementary, to potentially tackle a different set of problem domains while benefiting from  decades of developments in ILP theory. 

It is possible, of course, that a reasonable middle ground could be achieved by appealing to constraint logic programming \cite{jaffar1994constraint}, as seen in some early work \cite{Martin:1997:LLC:645325.649685,sebag1996constraint}. However, it is not entirely obvious such proposals capture the entire range of expressivity that one would with a SMT basis. In that regard, note that because we are not insisting on logic programming syntax, we are essentially motivating a reconsideration of inductive declarative programming, as opposed to purely inductive logic programming. 

It is also worth remarking that a number of recent developments relate to the motivation here, such as in the field of constraint learning \cite{de2018learning}, all of which could be leveraged to strengthen the theoretical and algorithmic foundations of the proposed framework. 
In \cite{kolb2018learning},  a heuristic approach to learn SMT formulas capturing positive-only examples is considered. In \cite{mocanupac+}, the implicit learning of SMT formulas is investigated via a PAC formulation, while also allowing for  noisy examples. 
The work in \cite{molina2018mixed,buefftractable} can be seen as learning weighted SMT formulas, albeit simple ones corresponding to the difference logic fragment over binary connectives.



\section{Classical Setup} 
\label{sec:classical_setup}

The basic concepts of logic programming are defined wrt a first-order language, where we have: {atoms} $p(t\sub 1, \ldots, t\sub n)$, consisting of  predicates \( p \) and terms $t\sub 1, \ldots, t\sub b$, understood as usual. 
Literals, clauses, definite clauses and grounding are understood as usual. 
Then, in the simplest instance, we have: \\[1ex]
\emph{Given a set of examples (or observations) \( E = \set{e\sub 1, \ldots, e\sub n} \), where \( e\sub i \) is a ground fact for the unknown target predicate \( p \), a background theory $B$ as a set of definition clauses, a space of  clauses \( \calL\sub h \) specified using a declarative bias, find a hypothesis \( H\subseteq \calL\sub h  \) where \( B \land H \models E. \)} \\[1ex]
The hypothesis would additionally need to satisfy certain rational generality properties, in the sense of maximally compressing \( E \) relative to \( B \) \cite{Muggleton:1991:ILP:104992.104994}. For example, consider an empty background theory with observations:   \emph{parent(f,c), parent(m,c), parent(g,f),grandparent(g,c)}, and \( \calL\sub h \) being the set of definite clauses. Then we may obtain the hypothesis:   \( grandparent(x,y) \leftarrow parent(x,z), parent(z,y) \).

	


\section{Revised Setup} 
\label{sec:revised_setup}


\textit{Satisfiability modulo theories} (SMT) is a generalization to SAT for deciding satisfiability for fragments and extensions of first-order logics with equality, for which specialized decision procedures have been developed. Deciding satisfiability for these logics is done with respect to some decidable background theory which fixes the interpretations of functions and predicates \cite{satisfiability-modulo-theories}. Briefly, we have: 


\textbf{Syntax} We assume a logical signature consisting of the set of predicates denoted as $\cal P$, and a set of functions symbols $\cal F$, including 0-ary functions,  logical variables, and standard connectives. 
An atom is one of the form: $b$ (a propositional symbol), $p(t_1,...,t_k)$, $t_1 = t_2$, $\bot$ (false), $\top$ (true). Literals and clauses are understood as usual.  A ground expression is one where all the variables are replaced by the domain of discourse  (e.g., integers, reals, finite set of named objects). 

\textbf{Semantics} Formulas are given a truth value from the set $\{true, false \}$ by means of first order models.  A model $\pmb{\rho}$ is a pair consisting of a non-empty set $\Sigma$, the universe of the model and a mapping assigning to each constant symbol $a$ an element $a \in \Sigma$ (the domain), to each function symbol $f \in \cal F$ of arity $n > 0$ a total function $f : \Sigma^n \rightarrow \Sigma$, to each propositional symbol $b$ an element $b \in \{true, false \}$ and to each predicate $p \in \cal P$ of arity $n > 0$ a total function $ p : \Sigma^n \rightarrow \{ true, false \}$. 
Terms are interpreted as usual, as is the satisfaction relation that is defined inductively.  We assume entailment wrt a suitable background theory (e.g., reals).   \smallskip 


\renewcommand{\partial}{\rho}
\newcommand{\full}{\pmb{\rho}}

A general setting for induction can be taken to be defined over the following languages \cite{Muggleton:1991:ILP:104992.104994}: \( \calL \sub e \) (the language of examples), \( \calL \sub b \) (the language for the background knowledge) and \( \calL \sub h \) (the language for the hypothesis), and as can be inferred from above, given \( B\subseteq \calL\sub b \) and \( E \subseteq \calL\sub e \), the task is to find \( H \subseteq \calL\sub h \), such that \( H \land B \models E. \)  As far as the background knowlege and hypothesis is concerned, for the SMT setting, we could now imagine \(  \calL \sub b, \calL \sub h   \) being fragments of linear real arithmetic, for example, but this is not necessary. \( \calL \sub b \) could involve inductive constraints, and both \( \calL \sub b \) and \( \calL\sub h \) could involve non-linear constraints. The search for the hypothesis could be achieved in the first instance by appealing to reductions of the induction step to  satisfiability~\cite{evans2018learning}.  So, not very much changes at first glance, which is a positive development for bridging existing ILP theory and frameworks with this new setting. Moreover, other entailment judgements for strengtening  ILP, e.g., that negative examples where provided should never be entailed by \( B \land H \)  could be applied here too.  Nonetheless, note that \(  \calL \sub b, \calL \sub h   \) are richer in some regards, and not restricted to Horn clauses. We will now  discuss some variants below for \( \calL \sub e.  \)  

\textbf{$\calL\sub e = $ partial models}
%
%
 In the simplest instance, we have observations that are partial models. For example, in a language with 0-ary functions \( \set{x,y, \ldots,z} \) over the reals, a full model  may be of the form \( (x=1, y=2.3, \ldots, z=6) \). In this case, an example might be a partial model  of the form \( (x=1) \), and another might be of the form \( (y=2.3) \). Clearly, if the conjunction of the partial models is taken as \( H \), then trivially \( B \land H \models E. \) But this is a not a very interesting hypothesis. Like in the classical setting, we would need to specify \( \calL\sub h \) in a way to maximally compress the examples wrt \( B \); so, for example, if \( B \) is \( x+y \gt z, \) we might infer \( H \) as \( y\gt x \) by specifying length/syntax  restrictions on \( H \) wrt \( E. \) 


\textbf{$\calL\sub e = $ sets of partial models} It is  natural to imagine that the observations are, in fact, sets of partial models. For example, if we were unable to measure \( x \) precisely, we may need to contend with \( x \gt 0 \). So, given examples \( x\gt 0  \) and \( z=0, \) we might infer the hypothesis \( x\gt z. \) 



\textbf{$\calL\sub e =$ $k$-ary functions} 
We have discussed the use of 0-ary functions above, and that is the case for much of the learning literature \cite{kolb2018learning,mocanupac+}. 
%
The natural analogue of the kind of examples seen in classical ILP  might better correspond to the use of \( k \)-ary functions with logical variables. 
Incidentally, this level of expressiveness is supported by SMT solvers \cite{satisfiability-modulo-theories} for capturing arrays, and so on. As an example, suppose \( B \) includes the expression \( {\it bmi}(x) = weight(x)/{height(x)}^2  \), and given examples (partial models)  \emph{weight(john) = 100, height(john) = 1.9,  weight(mary) = 70, height(mary) = 1.8,} we might  infer a hypothesis \( {\it bmi}(x) \gt 21. \)



\section{Conclusions} 
\label{sec:conclusions}

We motivated a reconsideration of inductive declarative programming by leveraging SMT. SMT solvers have  emerged to deftly handle arithmetic constraints and inductive definitions.  As a result, the languages that we could consider for \( B \) and \( H \) would be significantly richer so as to  capture a broad range of problems, and potentially impact the numerous  applications areas of SMT  \cite{satisfiability-modulo-theories}.  Of course, understanding how ideas from ILP systems  can be lifted for this methodology is an open but exciting  question. 
 Treating probabilistic concepts   \cite{probabilistic-inductive-logic-programming:} is another exciting direction, which would relate this framework to learning in hybrid probabilistic models \cite{probabilistic-inference-in-hybrid-domains} and statistical relational learning \cite{statistical-relational-ai:-logic-probability} more generally. It would allow one to express that each learned clause holds with a certain probability, but not categorically, for example. 





\bibliographystyle{aaai}

\end{document}

%% file: defnj.tex
\newtheorem{THEOREM}{Theorem}

                        {\end{THEOREM}}

\newtheorem{LEMMA}[THEOREM]{Lemma}
                      {\end{LEMMA}}
\newtheorem{COROLLARY}[THEOREM]{Corollary}
                          {\end{COROLLARY}}
\newtheorem{PROPOSITION}[THEOREM]{Proposition}
                            {\end{PROPOSITION}}
\newtheorem{DEFINITION}[THEOREM]{Definition}
                            {\end{DEFINITION}}
\newtheorem{CLAIM}[THEOREM]{Claim}
                            {\end{CLAIM}}
\newtheorem{EXAMPLE}[THEOREM]{Example}
                            { \end{EXAMPLE}}
\newtheorem{REMARK}[THEOREM]{Remark}
                            {\end{REMARK}}
							\newtheorem{NOTATION}[THEOREM]{Notation}
							                            {\end{NOTATION}}

\DeclareMathAlphabet{\mathitbf}{OML}{cmm}{b}{it}

\newcommand{\sub}{_}

\newcommand{\gt}{>}

\newcommand{\calL}{{\cal L}}

\newcommand{\set}[1]{\left\{ #1 \right\}}






%% file: smt-arxiv-corrected.bbl
\begin{thebibliography}{}

\bibitem[\protect\citeauthoryear{Barrett \bgroup et al\mbox.\egroup
  }{2009}]{satisfiability-modulo-theories}
Barrett, C.; Sebastiani, R.; Seshia, S.~A.; and Tinelli, C.
\newblock 2009.
\newblock Satisfiability modulo theories.
\newblock In {\em Handbook of Satisfiability}. IOS Press.
\newblock chapter~26,  825--885.

\bibitem[\protect\citeauthoryear{Belle, Passerini, and Van~den
  Broeck}{2015}]{probabilistic-inference-in-hybrid-domains}
Belle, V.; Passerini, A.; and Van~den Broeck, G.
\newblock 2015.
\newblock Probabilistic inference in hybrid domains by weighted model
  integration.
\newblock In {\em IJCAI}.

\bibitem[\protect\citeauthoryear{Bueff, Speichert, and
  Belle}{2018}]{buefftractable}
Bueff, A.; Speichert, S.; and Belle, V.
\newblock 2018.
\newblock Tractable querying and learning in hybrid domains via sum-product
  networks.
\newblock In {\em KR Workshop on Hybrid Reasoning and Learning}.

\bibitem[\protect\citeauthoryear{Chistikov, Dimitrova, and
  Majumdar}{2015}]{approximate-counting-in-smt-and-value-estimation}
Chistikov, D.; Dimitrova, R.; and Majumdar, R.
\newblock 2015.
\newblock Approximate counting in {SMT} and value estimation for probabilistic
  programs.
\newblock In {\em TACAS}, volume 9035.
\newblock  320--334.

\bibitem[\protect\citeauthoryear{De~Raedt \bgroup et al\mbox.\egroup
  }{2008}]{probabilistic-inductive-logic-programming:}
De~Raedt, L.; Frasconi, P.; Kersting, K.; and Muggleton, S., eds.
\newblock 2008.
\newblock {\em Probabilistic inductive logic programming: theory and
  applications}.
\newblock Berlin, Heidelberg: Springer-Verlag.

\bibitem[\protect\citeauthoryear{De~Raedt, Passerini, and
  Teso}{2018}]{de2018learning}
De~Raedt, L.; Passerini, A.; and Teso, S.
\newblock 2018.
\newblock Learning constraints from examples.
\newblock In {\em Thirty-Second AAAI Conference on Artificial Intelligence}.

\bibitem[\protect\citeauthoryear{Evans and
  Grefenstette}{2018}]{evans2018learning}
Evans, R., and Grefenstette, E.
\newblock 2018.
\newblock Learning explanatory rules from noisy data.
\newblock {\em Journal of Artificial Intelligence Research} 61:1--64.

\bibitem[\protect\citeauthoryear{Gao, Kong, and
  Clarke}{2013}]{dreal:-an-smt-solver-for-nonlinear-theories}
Gao, S.; Kong, S.; and Clarke, E.~M.
\newblock 2013.
\newblock dreal: An smt solver for nonlinear theories over the reals.
\newblock In {\em CADE},  208--214.

\bibitem[\protect\citeauthoryear{Jaffar and Maher}{1994}]{jaffar1994constraint}
Jaffar, J., and Maher, M.~J.
\newblock 1994.
\newblock Constraint logic programming: A survey.
\newblock {\em The journal of logic programming} 19:503--581.

\bibitem[\protect\citeauthoryear{Kersting, Natarajan, and
  Poole}{2011}]{statistical-relational-ai:-logic-probability}
Kersting, K.; Natarajan, S.; and Poole, D.
\newblock 2011.
\newblock Statistical relational {AI}: Logic, probability and computation.

\bibitem[\protect\citeauthoryear{Kolb \bgroup et al\mbox.\egroup
  }{2018}]{kolb2018learning}
Kolb, S.; Teso, S.; Passerini, A.; and De~Raedt, L.
\newblock 2018.
\newblock Learning smt (lra) constraints using smt solvers.
\newblock In {\em IJCAI},  2333--2340.

\bibitem[\protect\citeauthoryear{Martin and
  Vrain}{1997}]{Martin:1997:LLC:645325.649685}
Martin, L., and Vrain, C.
\newblock 1997.
\newblock Learning linear constraints in inductive logic programming.
\newblock In {\em Proceedings of the 9th European Conference on Machine
  Learning}, ECML '97,  162--169.
\newblock London, UK, UK: Springer-Verlag.

\bibitem[\protect\citeauthoryear{Mocanu, Belle, and Juba}{2019}]{mocanupac+}
Mocanu, I.~G.; Belle, V.; and Juba, B.
\newblock 2019.
\newblock Pac+ smt.
\newblock In {\em NeurIPS Workshop on Knowledge Representation \& Reasoning
  Meets Machine Learning}.

\bibitem[\protect\citeauthoryear{Molina \bgroup et al\mbox.\egroup
  }{2018}]{molina2018mixed}
Molina, A.; Vergari, A.; Di~Mauro, N.; Natarajan, S.; Esposito, F.; and
  Kersting, K.
\newblock 2018.
\newblock Mixed sum-product networks: A deep architecture for hybrid domains.
\newblock In {\em Thirty-second AAAI conference on artificial intelligence}.

\bibitem[\protect\citeauthoryear{Muggleton and
  De~Raedt}{1994}]{inductive-logic-programming:-theory}
Muggleton, S., and De~Raedt, L.
\newblock 1994.
\newblock Inductive logic programming: Theory and methods.
\newblock {\em The Journal of Logic Programming} 19:629--679.

\bibitem[\protect\citeauthoryear{Muggleton \bgroup et al\mbox.\egroup
  }{2012}]{muggleton2012ilp}
Muggleton, S.; De~Raedt, L.; Poole, D.; Bratko, I.; Flach, P.; Inoue, K.; and
  Srinivasan, A.
\newblock 2012.
\newblock Ilp turns 20.
\newblock {\em Machine learning} 86(1):3--23.

\bibitem[\protect\citeauthoryear{Muggleton}{1991}]{Muggleton:1991:ILP:104992.104994}
Muggleton, S.
\newblock 1991.
\newblock Inductive logic programming.
\newblock {\em New Gen. Comput.} 8(4):295--318.

\bibitem[\protect\citeauthoryear{Nitti}{2016}]{hybrid-probabilistic-logic-programming}
Nitti, D.
\newblock 2016.
\newblock {\em Hybrid Probabilistic Logic Programming}.
\newblock Ph.D. Dissertation, KU Leuven.

\bibitem[\protect\citeauthoryear{Reynolds and
  Kuncak}{2015}]{reynolds2015induction}
Reynolds, A., and Kuncak, V.
\newblock 2015.
\newblock Induction for smt solvers.
\newblock In {\em International Workshop on Verification, Model Checking, and
  Abstract Interpretation},  80--98.
\newblock Springer.

\bibitem[\protect\citeauthoryear{Sebag and
  Rouveirol}{1996}]{sebag1996constraint}
Sebag, M., and Rouveirol, C.
\newblock 1996.
\newblock Constraint inductive logic programming.

\bibitem[\protect\citeauthoryear{Speichert and
  Belle}{2018}]{speichert2018learning}
Speichert, S., and Belle, V.
\newblock 2018.
\newblock Learning probabilistic logic programs in continuous domains.

\end{thebibliography}
